# Motorcycle Classification in Urban Scenarios using Convolutional Neural Networks for Feature Extraction

**Jorge E. Espinosa[†], Sergio A. Velastin [††], and John W. Branch [†††]**

[†]Facultad de Ingenierías, Politécnico Colombiano Jaime Isaza Cadavid – Medellín –Colombia
[††] University Carlos III - Madrid Spain
[†††]Facultad de Minas, Universidad Nacional de Colombia – Sede Medellín
jeespinosa@elpoli.edu.co, sergio.velastin@ieee.org, jwbranch@unal.edu.co

## Abstract

This paper presents a motorcycle classification system for urban scenarios using Convolutional Neural Network (CNN). Significant results on image classification has been achieved using CNNs at the expense of a high computational cost for training with thousands or even millions of examples. Nevertheless, features can be extracted from CNNs already trained. In this work AlexNet, included in the framework CaffeNet, is used to extract features from frames taken on a real urban scenario. The extracted features from the CNN are used to train a support vector machine (SVM) classifier to discriminate motorcycles from other road users. The obtained results show a mean accuracy of 99.40% and 99.29% on a classification task of three and five classes respectively. Further experiments are performed on a validation set of images showing a satisfactory classification.

## 1 Introduction

Urban traffic analysis is crucial for traffic management. Traditionally, urban traffic analysis is carried using inductive loop sensor information and video. Video detection is a complex problem and under high vehicle densities frequent occlusions complicate the analysis. Motorcycles are becoming important and numerous road users. Therefore, motorcycle detection is becoming very important, specially in developing countries, where they are involved in a large share of the traffic and so high accident rates are reported, due to the high vulnerability of this kind of vehicle. Until now most motorcycle video detection systems are implemented using conventional feature extraction techniques such as 3D models [1] [2], vehicle dimensions [3] [4] [5], symmetry, color, shadow, geometrical features and texture and even wheel contours [6]. Other works report the use of stable features [7], HOG for evaluation of helmet presence [8] [9], variations of HOG [10] [11], and the use of SIFT, DSIFT and SURF [12].

On the other hand, *deep learning theory* (DL) applied to image processing has been a breakthrough in computer vision especially in tasks such as image recognition. Since 2010 an annual image recognition challenge known as the ImageNet Large-Scale Visual Recognition Competition (ILSVRC) [13] has been established, showing impressive results, even overcoming human classification skills. Given these results, DL is a promising technology for vehicle recognition on Intelligent Transportation Systems (ITS).

Deep learning strategies are successfully implemented for vehicle detection as 2D deep belief network (2D-DBN) [14], appearance-based vehicle type classification method [15], hybrid architecture (HDNN) which overcomes the issue of the single scale extraction features in DNNs [16], also color based features [17], convolutional binary classifier (CNNB) [18], and working on low-resolution images in [19]. Nevertheless, as far as we know, there are no reports of DL strategies used for motorcycle classification, nor on the use of CNNs already trained for feature extraction to perform vehicle discrimination.

This work uses an already trained CNN network for the task of classifying motorcycles images, first applied to static images and then extended to evaluate ambiguous images. The paper is organized as follows: section 2 gives a brief explanation of deep learning along with an overview of convolutional neural networks and the advantage of the use of an already-trained network for feature extraction. Section 3 shows the classification task, describing the advantages of using GPU architectures to accelerate the entire process. Section 4 shows the results of applying the strategy on different kind of images. Section 5 presents the conclusions and proposes some future work.

## 2 Deep Learning and CNN overview

Deep learning (DL) theory has revolutionized the field of computer vision. DL mimics the principles governing the information representation in mammal brains, allowing to design new network structures facing different tasks. For pattern recognition, the techniques have shown robustness in classifications tasks, being able to deal with different ranges of transformations or distortions such as noise, scale, rotation, displacement, illuminance variance, etc. [20]. In object recognition, CNN´s feature representation often outperforms man crafted features such as LBP, SURF, HOG [21] [22] [23] extensively used for object detection.

CNNs [22] [24] are an evolution of Multilayer perceptron networks especially suited for capturing local "spatial" patterns in data, such as in images and videos. They are a type of neural network inspired by the retina function of the human visual system. This network has a special architecture that exploits spatial relationships, reducing the number of parameters to learn and improving the performance of back-propagation training algorithms. CNNs require minimal preprocessing of data before being introduced to the network. The input is analyzed dividing it in small areas, where a filter is applied to derive information and features representation. This information is propagated through different layers of the network, applying different kinds of filters and obtaining salient features from the analysed data. This strategy allows obtaining features invariant to rotation, scale or shift since the receptive fields give to the neuron access to basic features such as corners or oriented edges.

**2.1 Pre-training CNN**

Since training Convolutional Neural Networks (CNN) is a difficult and time-consuming process, in this work we start with a CNN already trained on a large dataset, and then adapted to the current problem. A pre-trained network can be used for two purposes:

- Feature extraction: where a CNN is used to extract features from data (in this case images) and then use the learned features to train a different classifier, e.g., a support vector machine (SVM). (This is the approach of this paper).
- Transfer learning: Where a network already trained on a big dataset is retrained in the last few layers on a more compact data set.

The importance of using features from an already trained CNN network was remarked in the work of Razavian et al. in [25], where they showed that generic descriptors extracted from convolutional neural networks are very powerful. This work uses the OverFed [26] network to tackle a different recognition task and object classification. Moreover, they report superior results compared to the state-of-the-art algorithms.

## 3 Motorcycle Classification Approach

In this work, vehicles are classified on two categories: motorcycles and cars. The classifier is constructed using a multiclass linear SVM trained with features obtained from a pre-trained CNN. Those features have been extracted for a set of 80 images per category, including the "urbTree" category created from the urban traffic environment. This approach to image category classification is based on the work published by Matlab in "Image category classification using deep learning" [28] and follows the standard practice of training an off-the-shelf classifier, using features extracted from images. The difference here is that features are extracted using a pretrained CNN instead of using conventional image features such as HOG or SURF. The classifier trained using CNN features provides close to 100% accuracy, which is higher than the accuracy achieved using methods such as Bag of Features and SURF.

The sets of images categories have been created corresponding to images related to motorcycles, cars and urban environment related objects ("urbTree"). The images were taken from different angles and perspectives in urban traffic in Medellin City (Colombia). One example of each category and a selection of the two sets can be observed in Figure 1

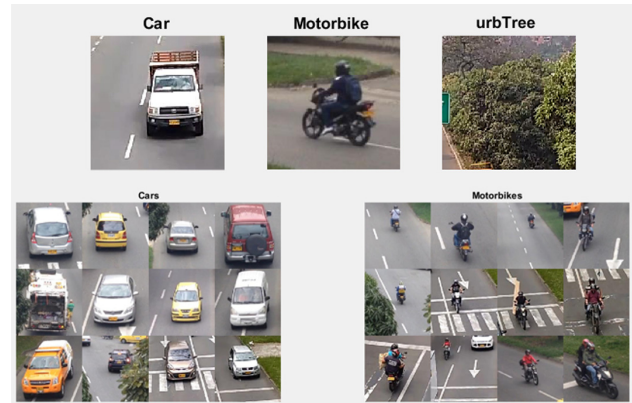

Figure 1 Three categories for classification, and examples of cars and motorbikes.

Using the selected images, a pre-trained CNN model is used for feature extraction. Pre-trained networks have gained popularity. Most of these have been trained on the ImageNet dataset [29]. We implement the classification by using the pre-trained network "AlexNet", from MatConvNet [21] [30]. The architecture of "AlexNet" has 23 layers. Integrating 5 convolution layers, 5 ReLu layers (Rectified units), 2 layers for normalization, 3 pooling layers, 3 fully connected layers, one probabilistic layer with softmax units and finally a classification layer ending in 100 neurons for 1000 categories.(Figure 2)

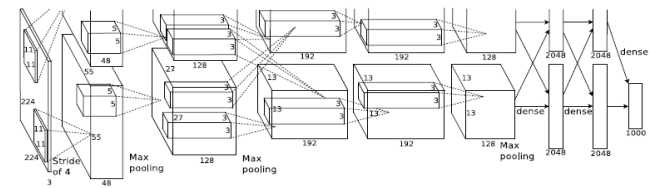

Figure 2 Alex Net Architecture. *"An illustration of the architecture of our CNN, explicitly showing the delineation of responsibilities between the two GPUs. One GPU runs the layer-parts at the top of the figure while the other runs the layer-parts at the bottom. The GPUs communicate only at certain layers. The network's input is 150,528-dimensional, and the number of neurons in the network's remaining layers is given by 253,440–186,624–64,896–64,896–43,264–4096–4096–1000."* [22]

The first layer is configured according to the dimensions of the input. This network is configured to receive images of 227 × 227 pixels, with RGB channels. That is 227 × 227 × 3. The first layer defines the input dimensions. Each CNN has a different input size requirements. The one used in this work ("AlexNet") requires image input that is 227 × 227 × 3.

The intermediate layers make up the bulk of the CNN. These are a series of convolutional layers, interspersed with rectified linear units (ReLU) and max-pooling layers [22]. Following these layers are 3 fully-connected layers.

The final layer is the classification layer and its properties depend on the classification task. Originally, the loaded CNN model was trained to solve a 1000-way classification problem, thus the classification layer has 1000 classes to identify. For this work, the AlexNet network is only used to classify three categories. This pre-trained network was used to learn motorcycles and cars features obtained from the extended dataset, with 80 images per category and 80 examples of the class "urbTree" created from the urban environment. At the end, the total number of examples is 240.

A simple preprocess step is done to prepare the analyzed images on an RGB 227 × 227 pixels. If the images come in grayscale, the gray channel should be replicated emulating the RGB channel requirement. There is not a preservation of the aspect ratio, nevertheless, the car and motorcycles classes are defined in the above resolution.

To avoid overfitting, the created data set is split into training and validation data. 30% of images are used as training data (72 images), and the remaining 70% (168 images), for validation. The split selection is randomized to avoid biasing the results. Since each layer of a CNN produces a response, or activation, the task is to identify which layer recover more informative features. As mentioned in [23], there are only a few layers within a CNN that can be used for image feature extraction. This work shows that the firsts layers of the network capture basic image features, such as edges, corners or blobs. Zeiler and Fergus [23] explain the power of convolution, and the information that can be found on each single layer of a given CNN network, from this it is understandable why the features extracted from deeper layers of a CNN are suitable for image recognition.

One of the strengths of these networks is the ability to combine "primitive" features as we go deep in the architecture, structuring high-level image features which involve more complexity. These high-level features, which have more information, are used on a recognition task, since they enclose a richer image representation [31]. We tested the different convolutional layers, and verified that the deeper ones have more information. For this work, the layer before the classification was selected to extract the features (Figure 3). This corresponds to the fully connected layer 7.

To extract the features, the training set is propagated through the network up to a specific layer, extracting activations responses to create a training set of features, which is used later for classification.

The activation process implies taking the end fully connected layer (fc7) and to evaluate every single training example through the network. In this case as result we have a vector of features with dimension 4096 × 72, corresponding to the activation obtained with every single example (24 examples per category) propagated through the network, until this fully connected layer.

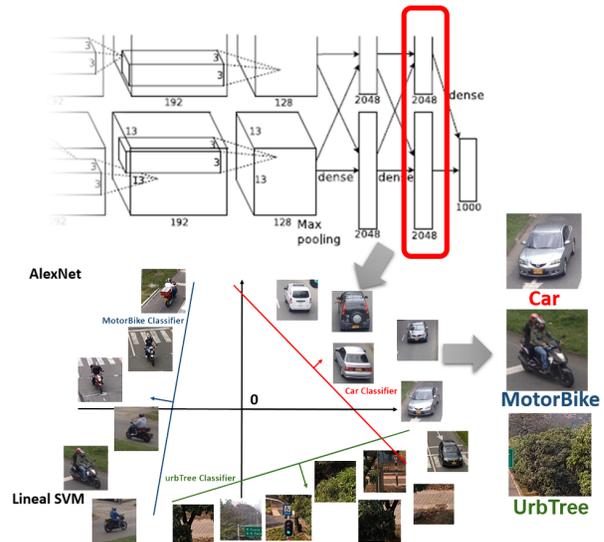

Figure 3 Training features are obtained from the fully connected layer (lying before the classification layer). Then the features are fed to a Linear SVM classifier.

### 3.1 GPU Use

Results in the ImageNet Large Scale Visual Recognition Challenge (ILSVRC) [13] have shown a dramatic improvement not just since the adoption of DL architectures, but also with implementation on GPU architectures, speeding up processing time and performance .

Because neural networks are created from large numbers of identical neurons, and the parameters that need to learn in the convolutional layers (where the filters are located) are highly parallel by nature, this parallelism can be exploited by GPUs, which provide a significant speed-up over CPU-only training. In a benchmarking using cuDNN with a CAFFE [32] neural network package, more than a 10-fold speed-up is obtained when training the "reference Imagenet" DNN model on an NVIDIA Tesla K40 GPU, compared to an Intel IvyBridge CPU [33].

In this work, the capabilities of CUDA GPU GeForce GT 750M are used for the features extraction. The time employed for the feature extraction is 1.95 seconds. Having 72 training examples (24 for each category) and using 4096 parameters of the fully connected layer. Using just CPU processing the time employed was 3.08 seconds.

### 3.2 Classification Stage

For the classification process, a multiclass SVM classifier is trained using the image features obtained from the CNN. Given the length of the feature vector (4096), the fast stochastic gradient descent solver was used as training algorithm. Note that the classifier is trained with only 72 examples (24 per category).

Once the classifier is trained, it is used to classify the validation set, which corresponds to the remaining 168 examples (56 by category). Classifier accuracy is evaluated now with the features obtained on this set. Figure 4 shows the results represented in a confusion matrix. The classifier only mismatches a car example, giving it a "Motorcycle" label. The mean accuracy obtained is 0.9940.

Figure 4 Confusion Matrix of the experiments.
(Class 1=Cars. 2 = Motorcycles. 3 = urbTree)

To mimimise any overfitting, cross validation process was implemented. Working with bins of 10 examples and evaluating in two strategies: 90% for training examples and 10% of evaluation with 100% of accuracy. After swapping the sets with 10% for training and 90% for evaluation, the mean accuracy obtained was 99.31%.

When features are obtained using different images classes but without re-training the SVM, the results show an accuracy of only 67%.

## 4  Experiments and Results

Given the behaviour obtained, the set is extended to deal with other two categories already presented in the Caltech dataset, this extension is made to evaluate the possible conflictive features that can be obtained in the experiment (Figure 5).

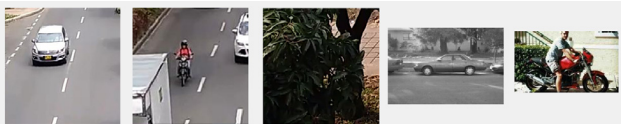

Figure 5 Experiment Set extension.
(Class 1=Cars, Class 2 = Motorcycles, Class 3 = urbTree, Class 4=Car side,   Class 5=Motorcycles side )

With this extension, the classification task includes five different categories, including two very related ones: cars and motorcycles extracted from the urban traffic. This images are obtained from side views and are already collected in the Caltech dataset. The training set corresponds to 24 examples for each category, the validation set on 56 examples for each category as well. The results are shown in the confusion matrix (Figure 6). In this case the classifier mismatches two "urbTree" examples, giving it the label of cars from the side. The mean accuracy obtained is 0.9929.

Figure 6 Confusion Matrix of the extended experiments.
(Class 1=Cars    2 = Motorcycles.  3 = urbTree    4 = cars side  5 = Motorcycles side)

Evaluating the classifier with a different set of images, the results can be sorted in decreasing order of confidence, or $P(y_i|x_i)$. The results describe the evaluation or confidence given to each class. (Figure 7 and Figure 8) Note the evaluation in ambiguity situations as in Figure 9.

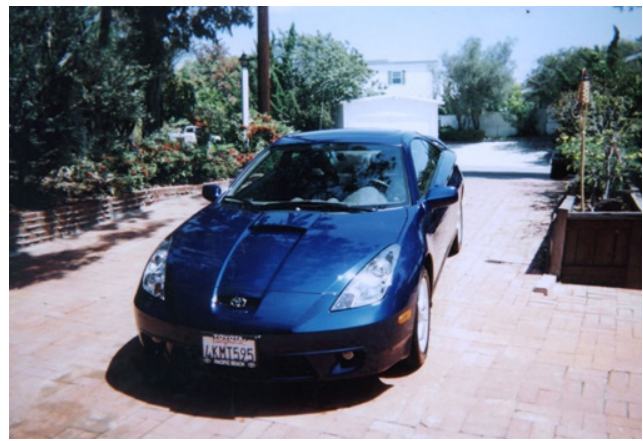

Figure 7 Classification results. (Source: Internet)
(Car 83 %     Motorcycle 49%    UbTree 50%)

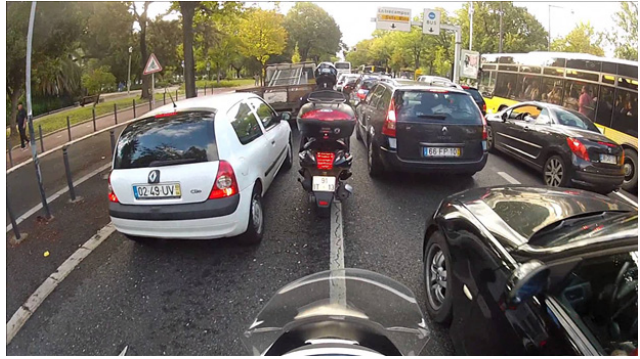

Figure 8 Classification results. (Source: Internet)
(Car 78.5%    Motorcycle 63.1%    urbTree 30%)

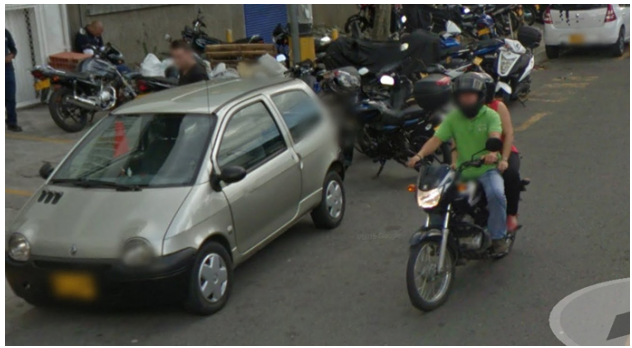

Figure 9 Classification results. (Source: Google Street View)
(Car 61%    Motorcycle 85%    urbTree 31%)

When the classifier is evaluated on a complete unrelated image the result describe a lower certainty on each class as is shown on Figure 10.

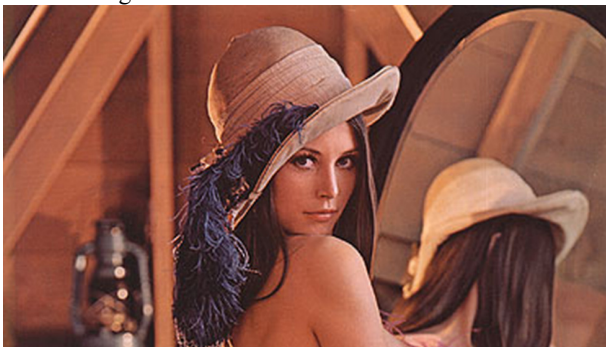

Figure 10 Unrelated Image. (Lena - Source: Internet)
(Car 34%    Motorcycle 21%    urbTree 32%)

## 5  Conclusions and Future Work

This paper has proposed the implementation of a motorbike classification scheme in urban scenarios using CNNs for feature extraction. The strategy takes already trained CNNs trained with millions of examples and being able to classify 1000 categories and uses its feature extraction capabilities to train a linear SVM for classifying three different classes. The feature extraction strategy is evaluated on different types of images, showing interesting results related to the classification confidence reached.

Future work will move toward the analysis of urban traffic videos, where detection and classification will be applied and enriched with a wider set of urban road user classes (e.g. trucks, vans, cyclists, pedestrians).

## Acknowledgements

S.A. Velastin is grateful to funding received from the Universidad Carlos III de Madrid, the European Union's Seventh Framework Programme for research, technological development and demonstration under grant agreement no. 600371, el Ministerio de Economía y Competitividad (COFUND2013-51509) and Banco Santander

## References

[1] S. Messelodi, C. M. Modena, and G. Cattoni, 'Vision-based bicycle/motorcycle classification', *Pattern Recognit. Lett.*, vol. 28, no. 13, pp. 1719–1726, 2007.

[2] N. Buch, J. Orwell, and S. A. Velastin, 'Urban road user detection and classification using 3D wire frame models', *IET Comput. Vis.*, vol. 4, no. 2, pp. 105–116, Jun. 2010.

[3] C.-C. Chiu, M.-Y. Ku, and H.-T. Chen, 'Motorcycle detection and tracking system with occlusion segmentation', in *Image Analysis for Multimedia Interactive Services, 2007. WIAMIS'07. Eighth International Workshop on*, 2007, pp. 32–32.

[4] M.-Y. Ku, C.-C. Chiu, H.-T. Chen, and S.-H. Hong, 'Visual motorcycle detection and tracking algorithms', *WSEAS Trans. Electron.*, pp. 121–131, 2008.

[5] Y. Dupuis, P. Subirats, and P. Vasseur, 'Robust image segmentation for overhead real time motorbike counting', in *2014 IEEE 17th International Conference on Intelligent Transportation Systems (ITSC)*, 2014, pp. 3070–3075.

[6] B. Duan, W. Liu, P. Fu, C. Yang, X. Wen, and H. Yuan, 'Real-time on-road vehicle and motorcycle detection using a single camera', in *Industrial Technology, 2009. ICIT 2009. IEEE International Conference on*, 2009, pp. 1–6.

[7] N. Kanhere, S. Birchfield, W. Sarasua, and S. Khoeini, 'Traffic monitoring of motorcycles during special events using video detection', *Transp. Res. Rec. J. Transp. Res. Board*, no. 2160, pp. 69–76, 2010.

[8] J. Chiverton, 'Helmet presence classification with motorcycle detection and tracking', *Intell. Transp. Syst. IET*, vol. 6, no. 3, pp. 259–269, 2012.

[9] R. R. e Silva, K. R. Aires, and R. de MS Veras, 'Detection of helmets on motorcyclists', *Multimed. Tools Appl.*, pp. 1–25, 2017.

[10] Z. Chen and T. Ellis, 'Multi-shape Descriptor Vehicle Classification for Urban Traffic', in *2011 International Conference on Digital Image Computing Techniques and Applications (DICTA)*, 2011, pp. 456–461.

[11] Z. Chen, T. Ellis, and S. A. Velastin, 'Vehicle detection, tracking and classification in urban traffic',

in *2012 15th International IEEE Conference on Intelligent Transportation Systems*, 2012, pp. 951–956.
[12] N. D. Thai, T. S. Le, N. Thoai, and K. Hamamoto, 'Learning bag of visual words for motorbike detection', in *2014 13th International Conference on Control Automation Robotics Vision (ICARCV)*, 2014, pp. 1045–1050.
[13] 'ImageNet Large Scale Visual Recognition Competition (ILSVRC)'. [Online]. Available: http://www.image-net.org/challenges/LSVRC/. [Accessed: 24-Oct-2016].
[14] H. Wang, Y. Cai, and L. Chen, 'A vehicle detection algorithm based on deep belief network', *Sci. World J.*, vol. 2014, 2014.
[15] Z. Dong, M. Pei, Y. He, T. Liu, Y. Dong, and Y. Jia, 'Vehicle Type Classification Using Unsupervised Convolutional Neural Network', in *2014 22nd International Conference on Pattern Recognition (ICPR)*, 2014, pp. 172–177.
[16] X. Chen, S. Xiang, C. L. Liu, and C. H. Pan, 'Vehicle Detection in Satellite Images by Hybrid Deep Convolutional Neural Networks', *IEEE Geosci. Remote Sens. Lett.*, vol. 11, no. 10, pp. 1797–1801, Oct. 2014.
[17] C. Hu, X. Bai, L. Qi, P. Chen, G. Xue, and L. Mei, 'Vehicle Color Recognition With Spatial Pyramid Deep Learning', *IEEE Trans. Intell. Transp. Syst.*, vol. 16, no. 5, pp. 2925–2934, Oct. 2015.
[18] F. Zhang, X. Xu, and Y. Qiao, 'Deep classification of vehicle makers and models: The effectiveness of pre-training and data enhancement', in *2015 IEEE International Conference on Robotics and Biomimetics (ROBIO)*, 2015, pp. 231–236.
[19] C. M. Bautista, C. A. Dy, M. I. Mañalac, R. A. Orbe, and M. Cordel, 'Convolutional neural network for vehicle detection in low resolution traffic videos', in *2016 IEEE Region 10 Symposium (TENSYMP)*, 2016, pp. 277–281.
[20] I. Arel, D. C. Rose, and T. P. Karnowski, 'Deep Machine Learning - A New Frontier in Artificial Intelligence Research [Research Frontier]', *IEEE Comput. Intell. Mag.*, vol. 5, no. 4, pp. 13–18, Nov. 2010.
[21] J. Deng, W. Dong, R. Socher, L. J. Li, K. Li, and L. Fei-Fei, 'ImageNet: A large-scale hierarchical image database', in *IEEE Conference on Computer Vision and Pattern Recognition, 2009. CVPR 2009*, 2009, pp. 248–255.
[22] A. Krizhevsky, I. Sutskever, and G. E. Hinton, 'Imagenet classification with deep convolutional neural networks', in *Advances in neural information processing systems*, 2012, pp. 1097–1105.
[23] M. D. Zeiler and R. Fergus, 'Visualizing and understanding convolutional networks', in *European Conference on Computer Vision*, 2014, pp. 818–833.
[24] H. Lee, R. Grosse, R. Ranganath, and A. Y. Ng, 'Convolutional deep belief networks for scalable unsupervised learning of hierarchical representations', in *Proceedings of the 26th annual international conference on machine learning*, 2009, pp. 609–616.
[25] A. S. Razavian, H. Azizpour, J. Sullivan, and S. Carlsson, 'CNN Features Off-the-Shelf: An Astounding Baseline for Recognition', in *2014 IEEE Conference on Computer Vision and Pattern Recognition Workshops*, 2014, pp. 512–519.
[26] P. Sermanet, D. Eigen, X. Zhang, M. Mathieu, R. Fergus, and Y. LeCun, 'Overfeat: Integrated recognition, localization and detection using convolutional networks', *ArXiv Prepr. ArXiv13126229*, 2013.
[27] L. Fei-Fei, R. Fergus, and P. Perona, 'Learning generative visual models from few training examples: An incremental bayesian approach tested on 101 object categories', *Comput. Vis. Image Underst.*, vol. 106, no. 1, pp. 59–70, 2007.
[28] 'Image Category Classification Using Deep Learning - MATLAB & Simulink Example'. [Online]. Available: https://www.mathworks.com/help/vision/examples/image-category-classification-using-deep-learning.html. [Accessed: 28-Feb-2017].
[29] 'Caffe | Deep Learning Framework'. [Online]. Available: http://caffe.berkeleyvision.org/. [Accessed: 05-Sep-2016].
[30] A. Vedaldi and K. Lenc, 'Matconvnet: Convolutional neural networks for matlab', in *Proceedings of the 23rd ACM international conference on Multimedia*, 2015, pp. 689–692.
[31] J. Donahue *et al.*, 'DeCAF: A Deep Convolutional Activation Feature for Generic Visual Recognition.', in *ICML*, 2014, pp. 647–655.
[32] Y. Jia *et al.*, 'Caffe: Convolutional architecture for fast feature embedding', in *Proceedings of the 22nd ACM international conference on Multimedia*, 2014, pp. 675–678.
[33] 'Accelerate Machine Learning with the cuDNN Deep Neural Network Library', *Parallel Forall*, 07-Sep-2014. [Online]. Available: https://devblogs.nvidia.com/parallelforall/accelerate-machine-learning-cudnn-deep-neural-network-library/. [Accessed: 09-Dec-2016].